\documentclass[letterpaper, 10 pt, conference]{ieeeconf}  % Comment this line out if you need a4paper

\IEEEoverridecommandlockouts                              % This command is only needed if
\usepackage{graphics} % for pdf, bitmapped graphics files
\usepackage{epsfig} % for postscript graphics files
\usepackage{amsmath} % assumes amsmath package installed
\usepackage{amssymb}  % assumes amsmath package installed
\usepackage{subcaption}
\usepackage{dblfloatfix}
\usepackage{booktabs}
\usepackage{cite}
\usepackage[
    colorlinks=true,
    citecolor=blue,
    linkcolor=blue,
    urlcolor=blue
]{hyperref}

\title{\LARGE \bf
AnalogDepth: Multi-view Geometry from FPV drones under Analog Video Transmission
}
\author{Pedro F. Proença and André Amorim \thanks{NOVA School of Science of Technology}}

\author{
André Amorim$^{1}$ and Pedro F. Proença$^{1}$
\thanks{$^{1}$NOVA School of Science and Technology, NOVA University Lisbon, Caparica, Portugal.}
}
\begin{document}

\maketitle
\thispagestyle{empty}
\pagestyle{empty}

%%%%%%%%%%%%%%%%%%%%%%%%%%%%%%%%%%%%%%%%%%%%%%%%%%%%%%%%%%%%%%%%%%%%%%%%%%%%%%%%
\begin{abstract}

Analog video transmission (VTX) remains widespread in FPV drones due to low latency, weight and low cost. However analog VTX suffers from complex spatially structured image degradation which differ fundamentally from digital image corruption (e.g. AWGN) used in standard training augmentation. This work shows that this type of noise severely degrades the accuracy of Depth Anything 3 (DA3), a state-of-the-art feed forward visual geometry foundation model.
\par
To address this gap, we present AnalogDepth, a parameter-efficient training pipeline that adapts DA3 to analog FPV imagery using student-teacher knowledge distillation with Low-Rank Adaptation (LoRA) injected into the DINOv2 backbone. Rather than synthesizing noise analytically, we build a noise bank from static FPV recordings under diverse conditions and compare real-noise injection against PSD-matched Gaussian synthesis and AWGN as baselines. Experiments on six real FPV flight sequences across three indoor scenes show that training with our noise bank consistently reduces per-frame depth RMSE and 3D reconstruction Chamfer distance compared to the pretrained DA3 baseline and both Gaussian noise variants. These results demonstrate that replicating the spatial structure of real analog transmission noise is critical for effective adaptation.
\end{abstract}

%%%%%%%%%%%%%%%%%%%%%%%%%%%%%%%%%%%%%%%%%%%%%%%%%%%%%%%%%%%%%%%%%%%%%%%%%%%%%%%%
\section{INTRODUCTION}

Recent advances in feed-forward visual geometry models using large-scale pretraining have enabled strong generalization across diverse scenes and cameras. However, extending these models to analog video transmission (VTX) imagery remains largely unexplored. \par
Despite the advances of digital VTX technology, analog VTX remains widespread in FPV drone applications, particularly in micro-drones where cost, power consumption, and transmission latency are critical. Analog VTX introduces complex, spatially structured noise arising from radio-frequency channel degradation, composite-video distortion, and synchronization errors that differ fundamentally from standard camera sensor noise.

As we demonstrate in this work, this structured noise severely impacts the accuracy of Depth Anything~3 (DA3) ~\cite{lin2025depthanything3} even when standard AWGN augmentation is applied. To bridge this domain gap, we introduce AnalogDepth, a student-teacher knowledge distillation strategy with Low-Rank Adaptation (LoRA) ~\cite{hu2021lora} that efficiently adapts the DINOv2 backbone of DA3 to analog FPV imagery, trained with real analog noise samples drawn from a noise bank collected during static FPV recordings. Our experiments on six real flight sequences show that this approach significantly improves both depth accuracy and 3D reconstruction quality. Our paper makes the following contributions:
\begin{itemize}
    \item A parameter-efficient training pipeline for adapting a visual geometry foundation model (DA3) to analog FPV imagery, using student-teacher knowledge distillation with LoRA, requiring no access to the original pretraining code or data.
    \item A characterization of analog FPV VTX noise and a noise bank construction methodology, together with a comparison of three noise injection strategies: real noise bank, PSD-matched Gaussian synthesis, and AWGN.
    \item An evaluation benchmark comprising real FPV flight sequences with pseudo ground-truth depth derived from RealSense-DA3 alignment, supporting both per-frame depth RMSE and 3D reconstruction Chamfer distance metrics.
\end{itemize}

The noise bank, the six analog FPV flight sequences with their RealSense recordings, and the training and evaluation code will be released publicly upon acceptance.

\begin{figure}[t]
    \centering

    % Row 1
    \includegraphics[width=0.325\linewidth]{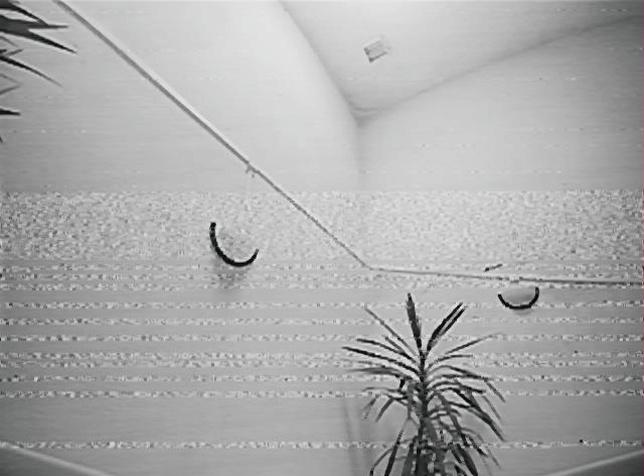}
    \includegraphics[width=0.325\linewidth]{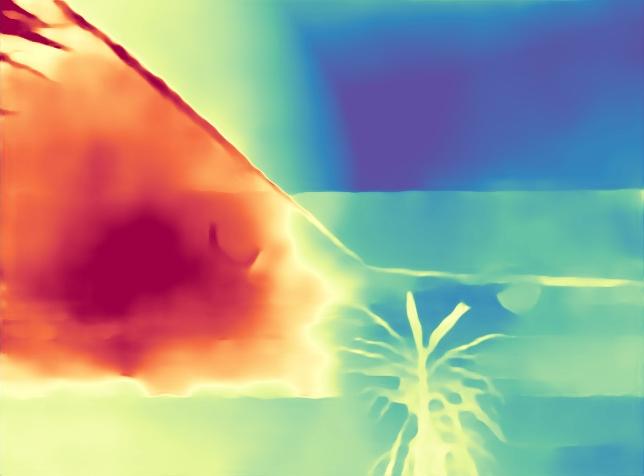}
    \includegraphics[width=0.325\linewidth]{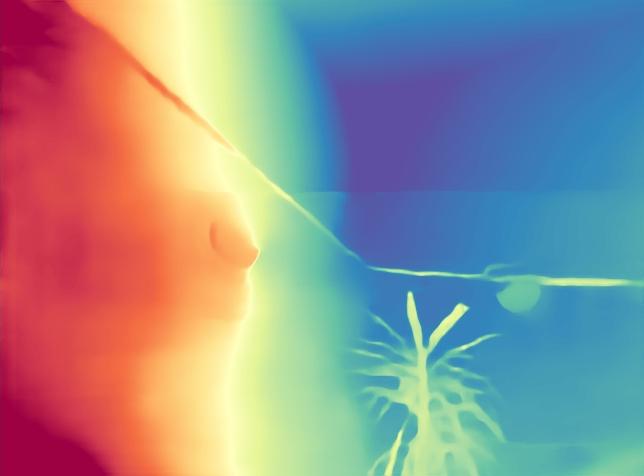}

    \vspace{2mm}

    % Row 2
    \includegraphics[width=0.325\linewidth]{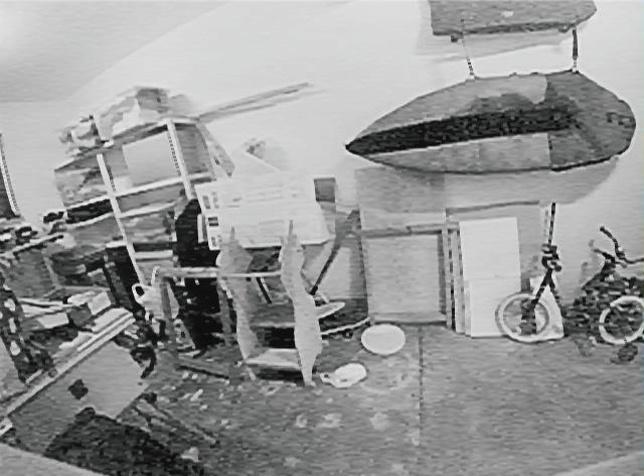}
    \includegraphics[width=0.325\linewidth]{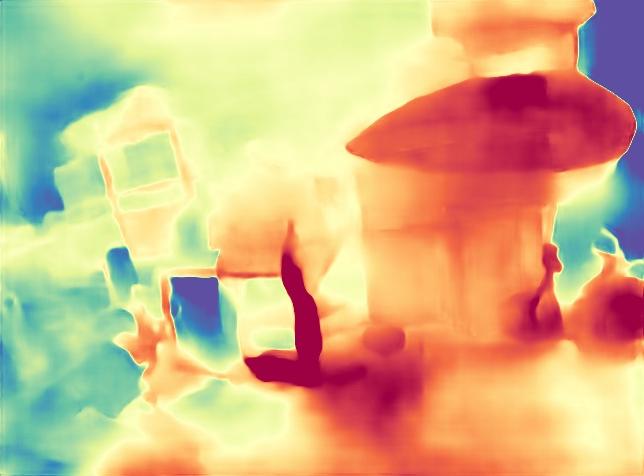}
    \includegraphics[width=0.325\linewidth]{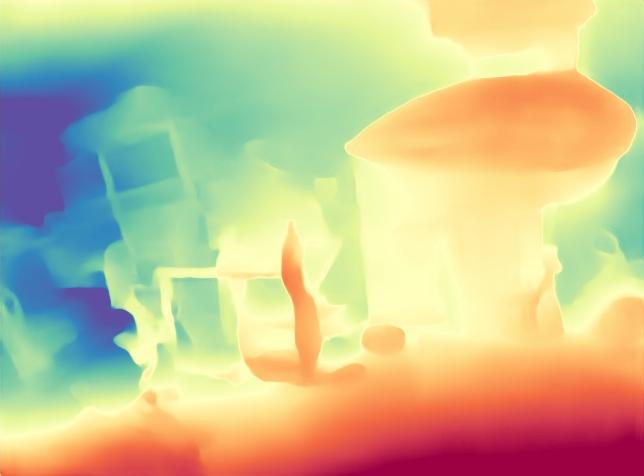}

        % Text row
    % \makebox[0.31\columnwidth]{Text 1}
     \makebox[0.31\columnwidth]{{\small VTX Analog Image}}
    \makebox[0.31\columnwidth]{{\small DA3 pretrained}}
    \makebox[0.31\columnwidth]{{\small AnalogDepth}}

    \caption{Depth estimation on real analog FPV imagery. (\textit{Top}) The official pretrained DA3-Small loses the wall gradient due to Horizontal noise bands, which are picked up as structure. AnalogDepth shows better wall gradient. (\textit{Bottom}) Even under milder corruption, DA3-pretrained distorts the floor geometry.}
    \label{fig:six_figures}
\end{figure}

\section{Related Work}

\textbf{Visual geometry foundation models.}
Recent models increasingly unify depth and 3D reconstruction within large pretrained visual backbones. Depth Anything V2~\cite{yang2024depthanythingv2} achieves strong zero-shot monocular depth estimation using large-scale pretrained features. DUSt3R~\cite{wang2024dust3r} and MASt3R~\cite{leroy2024mast3r} extend feed-forward inference to multi-view geometry, jointly recovering scene structure and camera parameters through local feature matching grounded in 3D. DepthPro~\cite{bochkovskii2024depthpro} achieves fast single-image metric depth estimation at high resolution, while VGGT~\cite{wang2025vggt} unifies multi-view scene reconstruction via a visual geometry grounded transformer. Depth Anything 3 (DA3)~\cite{lin2025depthanything3} further unifies monocular and multi-view geometry through a dense depth--ray representation predicted by a pretrained vision Transformer. These models exhibit strong cross-domain generalization, but are primarily developed and evaluated on conventional digital imagery and their robustness to analog video degradation remains largely unexplored.

\textbf{Robust depth under image degradation.}
Robustness to visual degradation is distinct from clean-domain generalization. RoboDepth~\cite{kong2023robodepth} shows that monocular depth estimators can degrade substantially under noise, blur, weather, and lossy compression (JPEG), motivating augmentation and corruption-aware training.

However, commonly used corruptions such as additive white Gaussian noise (AWGN) or generic blur do not capture the structured artifacts produced by analog FPV transmission, which may include spatially correlated interference, line artifacts, chromatic distortion, and transient signal degradation. For more complex degradation, monocular depth models were trained with diffusion model generation as input in \cite{gasperini2023robust} and \cite{jiang2025always} to improve performance under adverse weather and night.
In this work, we instead derive an empirical corruption bank from static recordings of a real analog FPV imaging and transmission pipeline and compare it directly with AWGN and power-spectral-density-matched noise.

\textbf{Image restoration and efficient adaptation.}
Modern restoration networks such as Restormer~\cite{zamir2022restormer}, NAFNet~\cite{chen2022nafnet}, and PromptIR~\cite{potlapalli2023promptir} achieve strong denoising and blind restoration performance. These methods aim to reconstruct a clean image, but improved perceptual quality does not necessarily preserve geometric cues required for depth estimation. Rather than inserting a separate restoration stage, we directly adapt DA3 to corrupted observations. To retain the pretrained geometric prior while limiting trainable parameters, we use Low-Rank Adaptation (LoRA)~\cite{hu2021lora} and train the model with corruption samples drawn from the measured analog FPV bank. This directly optimizes robustness for downstream depth estimation instead of image appearance.

\begin{figure}[t]
  \centering
  \includegraphics[width=\linewidth]{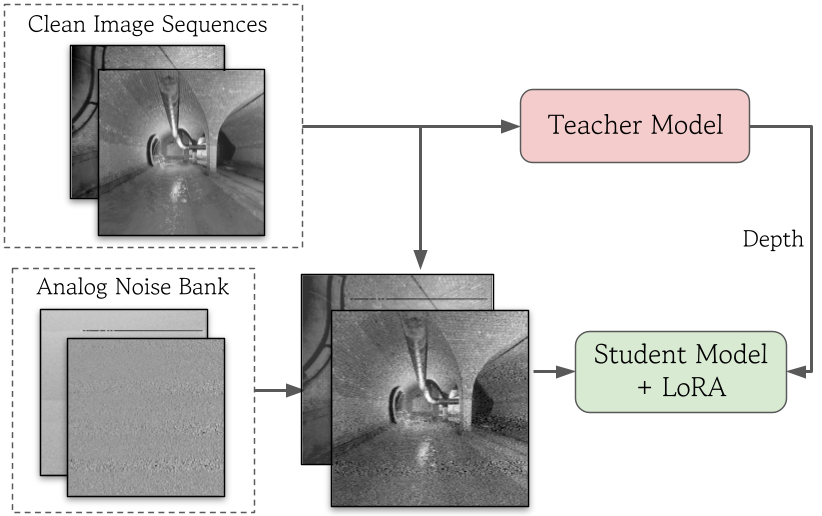}
  \caption{Training pipeline overview. One DA3 model used as teacher, with access to the clean images provides depth maps as supervision to the student model, which only sees the noisy images. The noisy images are composed by adding sampled analog VTX noise to the clean image sequence.}
  \label{fig:training_pipeline}
\end{figure}

\section{AnalogDepth}
\subsection{Training Pipeline}
Our training pipeline, shown in Fig . \ref{fig:training_pipeline},
follows a teacher-student Knowledge distillation scheme, where one frozen DA3 model, acting as the teacher, outputs $N$ depth maps from a sequence of $N$ clean images, sampled from our training set videos. These depth maps are used as supervision to train another DA3 model, which takes as input the same $N$ image views but perturbed with noise from our noise model.
\par
For both the teacher and student, we used simply two identical copies of DA3-Small since our goal is not compression. To train the student, LoRA ~\cite{hu2021lora}  parameters were injected in every attention block of the DINOv2 backbone and the student model base weights were frozen. This design choice was driven by early experiments where finetuning directly the backbone did not generalize to the validation set, and finetuning only the DA3 head failed to converge. This is intuitive since analog VTX noise should corrupt the local patch embeddings in lower layers and the structural feature representation in mid layers. More details of training are provided in Section \ref{sec:details}.
\par
To inject noise during training, we simply sample a noise image, using the methods proposed below, and add it to each clean image. In the case of noise bank samples, these are first center-cropped and downsized to match the input tensor. The same preprocessing is applied to our clean images following \cite{lin2025depthanything3}. Additionally, the camera intrinsics from the clean image sequences are adjusted to account for the image transformation and passed as inputs to the DA3.

\begin{figure}[t]
    \centering

    % Row 1
    \includegraphics[width=0.325\linewidth]{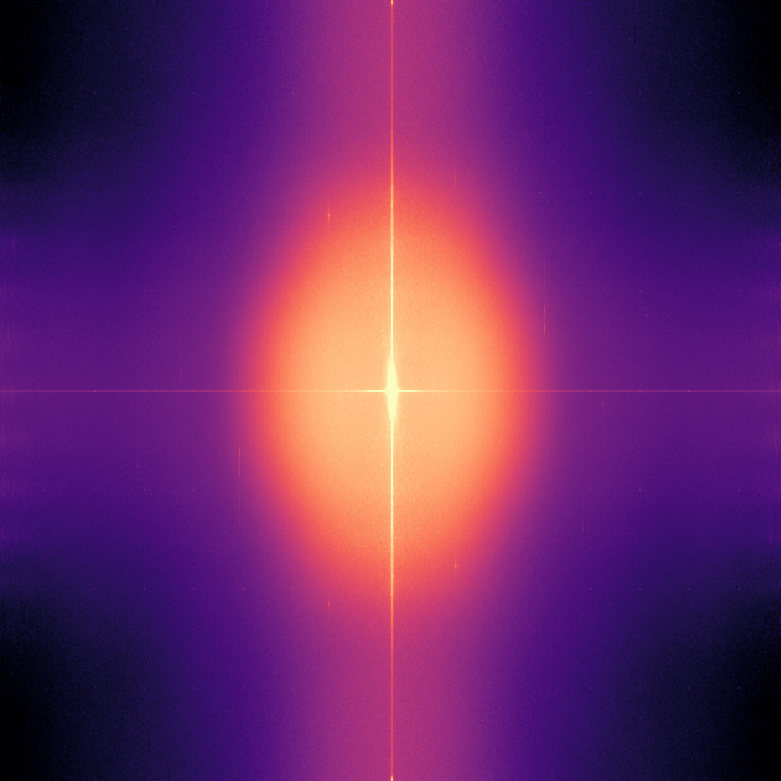}
    \includegraphics[width=0.325\linewidth]{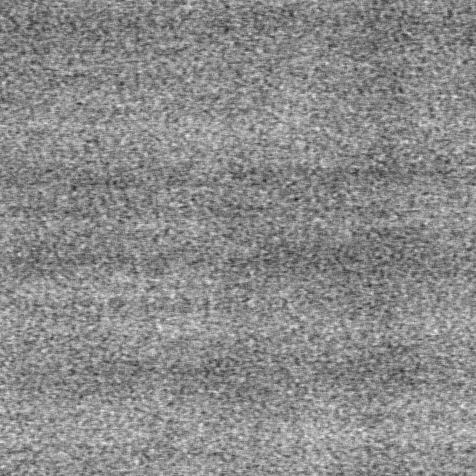}
    \includegraphics[width=0.325\linewidth]{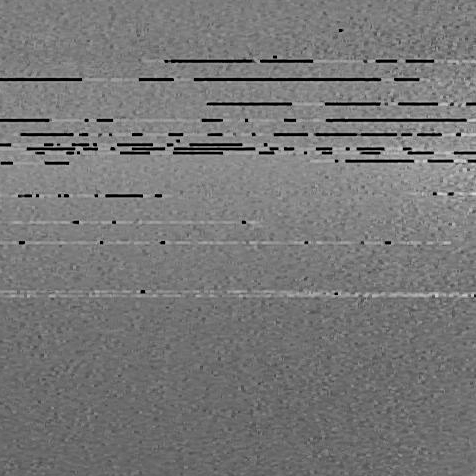}
    \caption{ (\textit{Left}) Average PSD for all image residuals used in the noise bank. (\textit{Middle}) Noise synthesized from the PSD. (\textit{Right}) Noise bank residual.}
    \label{fig:noise_analysis}
\end{figure}

\subsection{Analog VTX Noise Extraction}
\label{sec:noise}
To extract analog VTX noise, we collected several static videos across different lighting and scene conditions, e.g., placing metallic objects between the receiver and transmitter. Our VTX setup is described in Section \ref{sec:analogflights}. \par

For each video, we computed residual maps from each frame to the median of all images in each video. To avoid scene leaking into the residuals, we found necessary to clamp the drone with its FPV camera facing a textureless surface without sharp gradients.
The Power Spectral Density (PSD) averaged for all residual maps is shown in Fig. \ref{fig:noise_analysis} along with some sampled image noise. Our analysis of the power spectrum revealed: (i) a spectral flatness of 0.218, (ii) most energy concentrated at low frequencies, (iii) strong banding expressed as a clear vertical axis, and (iv) a strong DC component arising from full image brightness flicker. Therefore, we can conclude the noise cannot be approximated with AWGN. We can instead draw noise via spectral synthesis~\cite{kasdin1995} as shown in Fig. \ref{fig:noise_analysis} to approximate the noise's spectral characteristics but it does not replicate the sharp band artifacts.

Nevertheless, we have decided to test drawing noise during training using three approaches: AWGN, Noise Bank and PSD-matching Gaussian noise. For AWGN, we match the variance to the noise bank variance. For the noise bank, we simply store the residual maps and randomly sample during training, while the train and validation residuals are kept strictly disjoint. For the PSD-matching Gaussian noise, we synthesize a new noise map per sample by filtering white Gaussian noise in the frequency domain so its expected power spectrum matches the pooled power spectrum from the residual maps.

\section{Experimental Setup}
\subsection{Training Data and Implementation Details}
\label{sec:details}

For clean training-set images, we sampled 44 scenes from the TartanAir2 \cite{wang2020tartanair} with a total of 66,078 images. These were further split for validation. For the noise bank, we collected 11,412 residual maps from 6 videos. These were strictly split in train and validation set.
\par
\begin{figure}[t]
  \centering
  \includegraphics[width=0.8\linewidth]{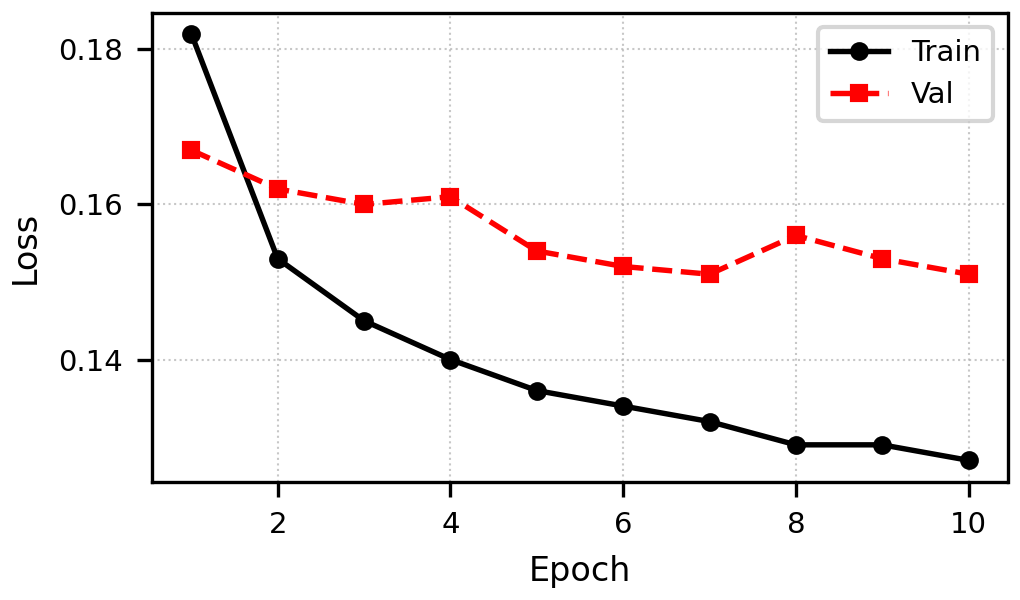}
  \caption{Total loss curve during training with noise bank}
  \label{fig:loss_curves}
\end{figure}

We trained DA3-Small model for 10 epochs resulting in the loss shown in Fig. \ref{fig:loss_curves}. The total loss simply combined the L1 depth loss $L_D$ with spatial gradient regularization $L_G$ proposed in~\cite{lin2025depthanything3} such that $L_{total} = L_D + \alpha L_G$ with $\alpha=1$. Since TartanAir2 is synthetic, we feed the models the extrinsics and intrinsics but these are not used directly in the loss function for simplicity since we didn't optimize the DA3 heads. Training optimized only the LoRA parameters injected in the backbone (i.e. DINOv2). \par

For data loading, we used a batch of 4, each with a sequence of 3 images. These images were sampled every 5th frame from each scene sequence to create parallax. The input image was fixed at 476 $\times$ 476. Thus the clean images and noise images were center-cropped and downsized. The full training took about 8 hours in a GeForce RTX 2080 Ti.

\subsection{Test Set: Analog FPV Flights}
\label{sec:analogflights}

We collected 6 video sequences around 1 min each from 3 indoor scenes using two Meteor75 Tiny FPVs using C03 Cameras with 128° FOV. We used a custom 5.8 GHz FM video receiver and recorded videos at 30 fps using a lossless intra-frame codec (FFV1). We then sampled and retained every 5th frame for our test-set. To obtain pseudo ground-truth depth, we collected RGB-D sequences in all scenes with a RealSense D455 from viewpoints approximating the FPV perspective.

\begin{figure*}[!tb]
    \centering

    % Row 1
    \includegraphics[width=0.235\textwidth]{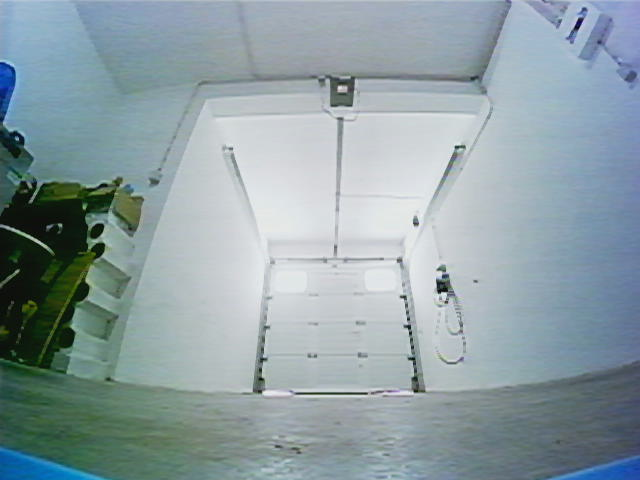} \hfill
    \includegraphics[width=0.235\textwidth]{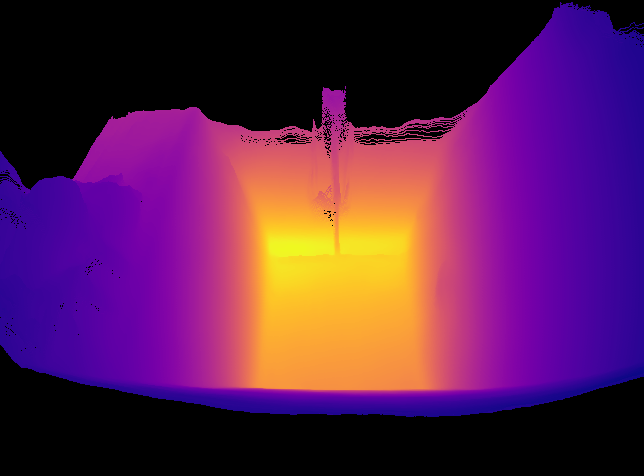} \hfill
    \includegraphics[width=0.235\textwidth]{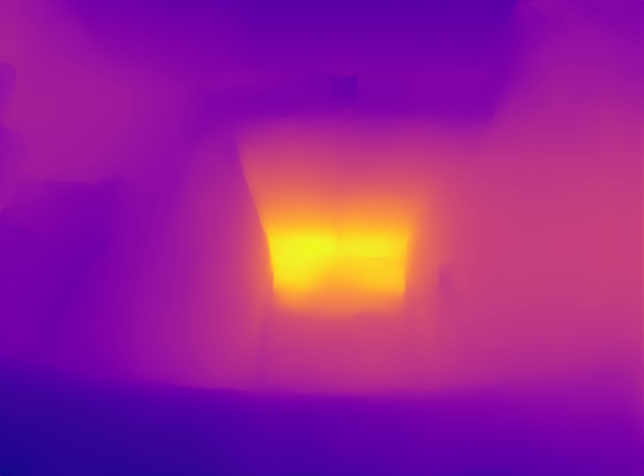} \hfill
    \includegraphics[width=0.235\textwidth]{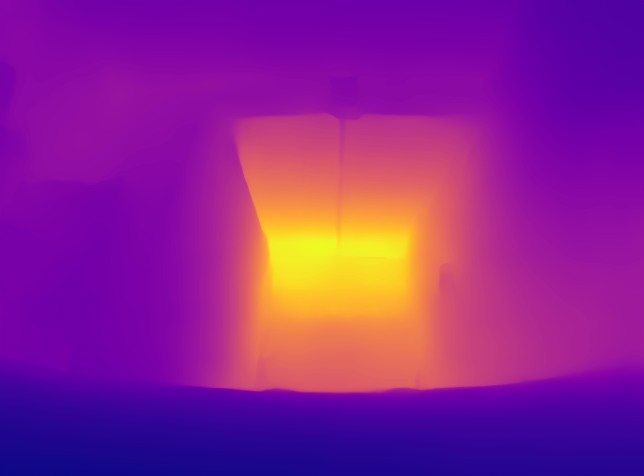}

    \vspace{2mm}

    % % Row 2
    \includegraphics[width=0.235\textwidth]{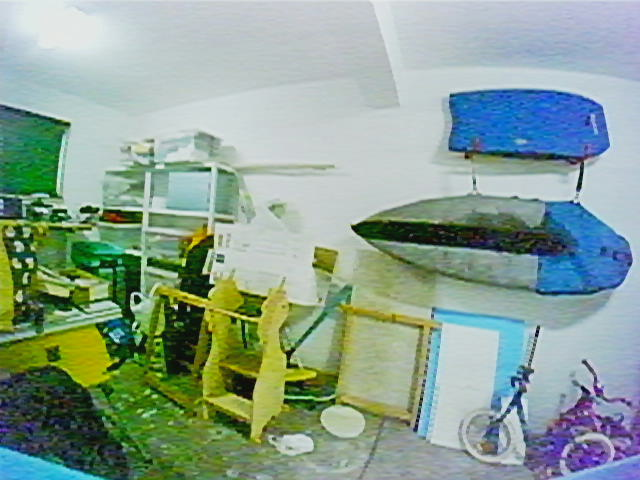} \hfill
    \includegraphics[width=0.235\textwidth]{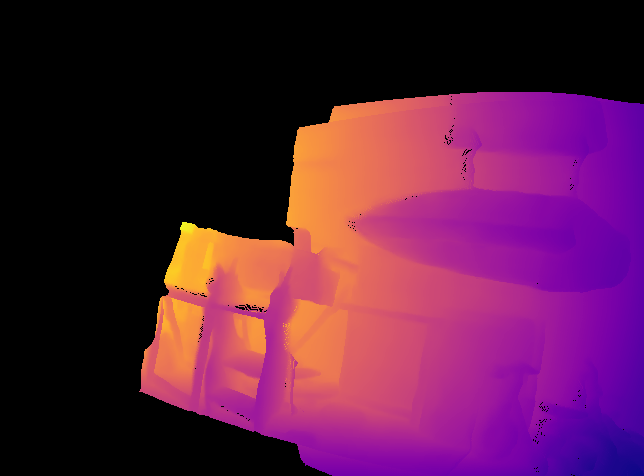} \hfill
    \includegraphics[width=0.235\textwidth]{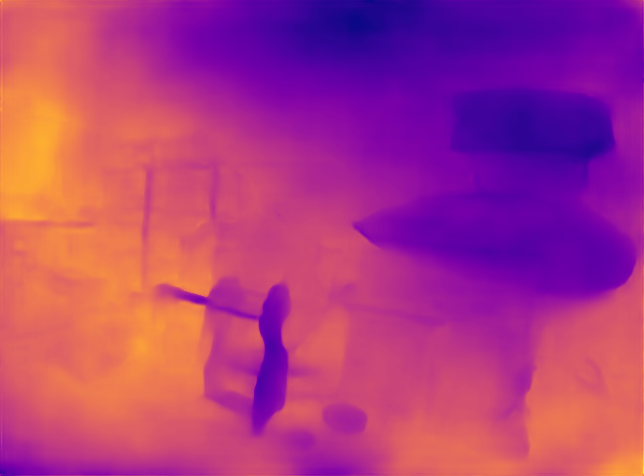} \hfill
    \includegraphics[width=0.235\textwidth]{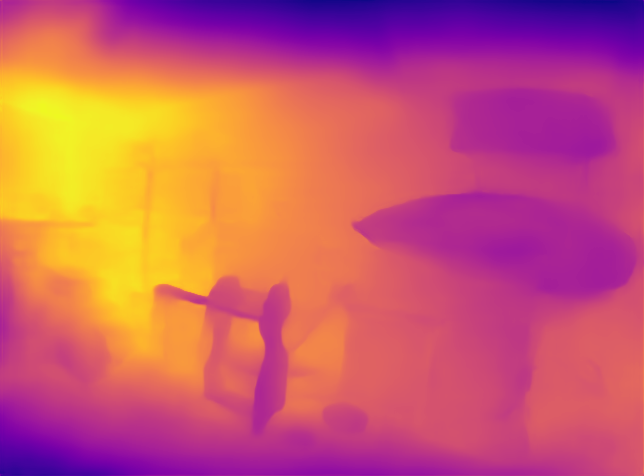}

    \vspace{2mm}

    % % Row 3
    \includegraphics[width=0.235\textwidth]{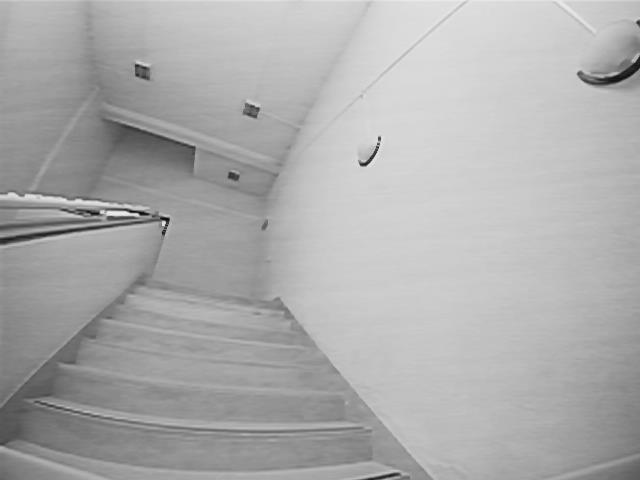} \hfill
    \includegraphics[width=0.235\textwidth]{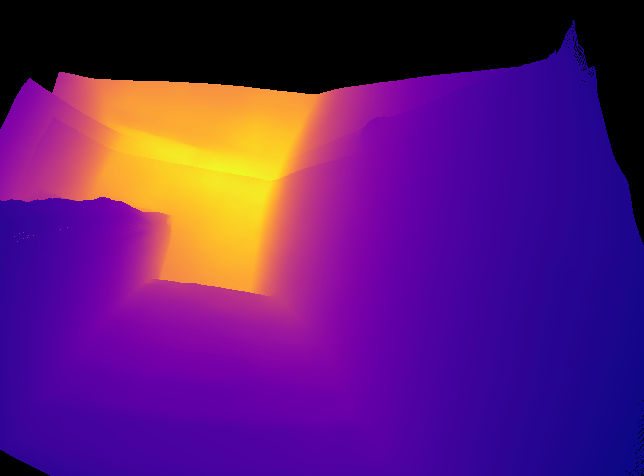} \hfill
    \includegraphics[width=0.235\textwidth]{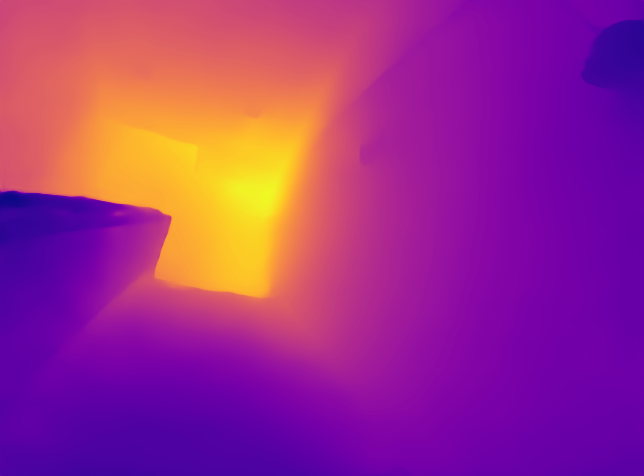} \hfill
    \includegraphics[width=0.235\textwidth]{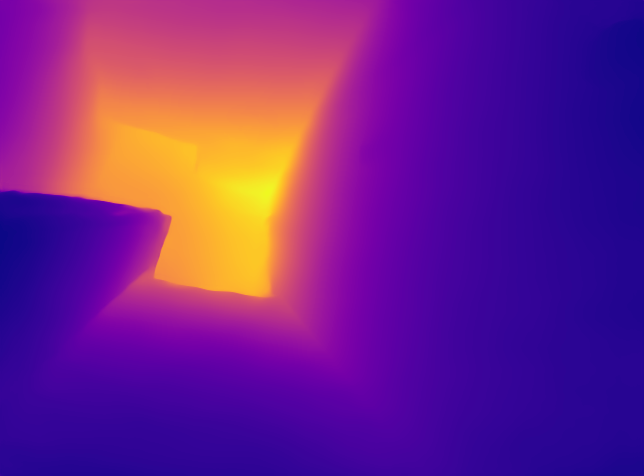}

    \makebox[0.235\textwidth]{{\small VTX Analog Image}}
    \makebox[0.235\textwidth]{{\small RealSense-DA3 (GT)}}
    \makebox[0.235\textwidth]{{\small DA3 pretrained}}
    \makebox[0.235\textwidth]{{\small AnalogDepth}}

    \caption{Depth maps for several flight sequence frames used in our evaluation. Notice the depth map for top image is less crisp for the DA3 pretrained. The right wall also shows less depth gradient for top and bottom image.}
    \label{fig:depth_mosaic}
\end{figure*}

\subsection{Evaluation Protocol and Metrics}

We evaluated the accuracy of both the depth-maps and the full 3D reconstructions from the different DA3 trained variants. To determine a suitable pseudo ground truth (GT), we first observed empirically in Fig. \ref{fig:depth_mosaic} and \ref{fig:3D_render} that using DA3-pretrained on the RealSense RGB frames produces cleaner depth maps and more consistent 3D reconstructions than the RealSense depth maps. This is consistent with the known depth error model \cite{proencca2018probabilistic} of these stereo cameras for longer ranges and depth discontinuities, while DA3 can leverage multiple views.
\par

We therefore decided to evaluate the models against two different ground-truth geometry references: The raw RealSense depth and the \emph{RealSense-DA3}. We constructed the \emph{RealSense-DA3} by running DA3-Small with its original pretrained weights on the RealSense RGB sequences and subsequently scaling the predicted depth maps using the corresponding RealSense depth maps. Importantly, this reference is generated using the original pretrained DA3 weights rather than AnalogDepth. This makes the evaluation conservative with respect to our method: any model-specific bias introduced by the DA3-based reference would be expected to favor the pretrained DA3 baseline rather than our AnalogDepth model. For best results and deriving a consistent 3D reconstruction, we used a maximum sequence of 100 frames per inference, which fits the GPU memory thanks to FlashAttention \cite{dao2023flashattention2}.
The evaluation against these two pseudo-GT references is shown in Tables \ref{tab:rmse} and \ref{tab:rmse_RealSense}. In the rest of the evaluation we used \emph{RealSense-DA3} due to reasons described above.

\textbf{Depth Map Accuracy.}
We evaluate the depth estimation accuracy in our test-set as the scale-aligned RMSE per frame:
\begin{equation}
  \mathrm{Depth\_RMSE} = \sqrt{
      \frac{1}{|\mathcal{V}|} \sum_{i \in \mathcal{V}}
      \left( D_i^{\text{reproj}} - s \cdot D_i^{\text{est}} \right)^2
  }
\end{equation}
where $\mathcal{V}$ is the set of all valid pixels. $D^{\text{est}}$ is the estimated DA3 depth from the analog image and the $D^{\text{reproj}}$ is the projected GT depth map and $s$ is the median per-pixel ratio over $\mathcal{V}$. We obtain the $D^{\text{reproj}}$ as follows. \par

For each test-set frame, we brute-force matched the image against the RealSense frames using LightGlue~\cite{lindenberger2023lightglue} feature matching. Then, we used RANSAC PnP with 3D cordinates from either the \emph{RealSense-DA3} or Realsense. We selected the top 3 matched images with an inlier threshold of 100 matches and reprojected the corresponding GT depth to the test-set image. Concretely, the depth maps from these Top-3 frames are reprojected and accumulated on the target frame using both camera intrinsics. To reduce the computational cost of this brute-force process, we sampled uniformly frames from both image sets. The actual number of frames that pass the RANSAC criteria and were evaluated is listed on Table \ref{tab:rmse} and \ref{tab:rmse_RealSense}.
\vspace{2mm}

\begin{table}[!tb]
  \centering
  \setlength{\tabcolsep}{5pt}
  \renewcommand{\arraystretch}{1.3}
  \begin{tabular}{lcccc}
  \toprule
  Video / nr\_frames & Pretrained & Noise bank & AWGN & PSD-matched \\
  \midrule
   garage\_west / 49 & 0.399 & \textbf{0.275} & 0.414 & 0.339 \\
  garage\_south / 46 & 0.297 & \textbf{0.273} & 0.291 & 0.284 \\
  office\_1 / 24 & 0.257 & \textbf{0.218} & 0.254 & 0.248 \\
  office\_2 / 46 & 0.342 & \textbf{0.299} & 0.332 & 0.317 \\
  plants / 15 & 0.502 & \textbf{0.275} & 0.497 & 0.419 \\
  stairs / 30 & 0.631 & \textbf{0.490} & 0.640 & 0.573 \\
  \bottomrule
  \end{tabular}
  \caption{Depth RMSE by scene and noise model using the RealSense-DA3 as reference. Each inference pass took 10 images. \textit{nr\_frames} is the number of frames evaluated that were selected by PnP protocol.}
  \label{tab:rmse}
  \end{table}

 \begin{table}[!tb]
  \centering
  \setlength{\tabcolsep}{5pt}
  \renewcommand{\arraystretch}{1.3}
  \begin{tabular}{lcccc}
  \toprule
  Video / nr\_frames & Pretrained & Noise bank & AWGN & PSD-matched \\
  \midrule
garage\_west / 49 & 0.422 & \textbf{0.295} & 0.440 & 0.357 \\
  garage\_south / 46 & 0.343 & \textbf{0.311} & 0.338 & 0.325 \\
   office\_1  / 20  $\dagger$& 1.077 & \textbf{1.056} & 1.076 & 1.073 \\
  office\_2 / 43 $\dagger$ & 1.970 & \textbf{1.967} & 1.969 & 1.968 \\
  plants / 15 & 0.785 & \textbf{0.572} & 0.781 & 0.703 \\
  stairs / 31 & 0.828 & \textbf{0.673} & 0.845 & 0.768 \\
  \bottomrule
  \end{tabular}
\caption{Depth RMSE by scene and noise model as in Table \ref{tab:rmse} but using RealSense depth as ground-truth reference. $\dagger$ RMSE is significantly higher for the office scene due to RealSense noise and outliers. }
\label{tab:rmse_RealSense}
\end{table}

\textbf{3D Reconstruction Accuracy.} We compared the point clouds generated by RealSense-DA3 to each test-set point cloud using the Chamfer Distance~\cite{fan2017pointcloud}. Each test-set point cloud was produced by running either DA3-pretrained or AnalogDepth on 100 uniformly sampled frames from each flight sequence.

Instead of relying on ICP for alignment, the point clouds were aligned using Umeyama Sim(3) closed-form solution on points matches found using the PnP RANSAC strategy above. More specifically, we aggregated 3D-to-3D matches from the inliers of the top 5 frame matches. The 3D coordinate of each keypoint was calculated by backprojecting it using the respective DA3 camera pose and depth map.

\subsection{Noise Model Results}

As shown in Table \ref{tab:rmse}, our training with a noise bank injection achieved consistently the best depth RMSE across all test frames. The gap between DA3-pretrained and AnalogDepth is preserved when we changed the pseudo ground truth to raw RealSense Depth, except in the office sequences, where  RealSense is too noisy. Both training with AWGN and PSD-matched Gaussian noise did not work as well. This indicates that the spatial pattern of noise matters and this is not captured by the average PSD where noise is drawn by varying only phase. It's also worth pointing that the loss gap in Fig. \ref{fig:loss_curves} was not as clear for the AWGN and PSD, thus expanding the noise bank could improve the results further. \par

Fig. \ref{fig:depth_mosaic} shows some examples of these evaluated frames. We can observe that even with comparatively low noise, DA3-pretrained fails systematically on textureless walls. This is clear in the reconstructed point clouds in Fig. \ref{fig:3D_render}.

\begin{figure}[!tb]
  \centering
  \includegraphics[width=0.8\linewidth]{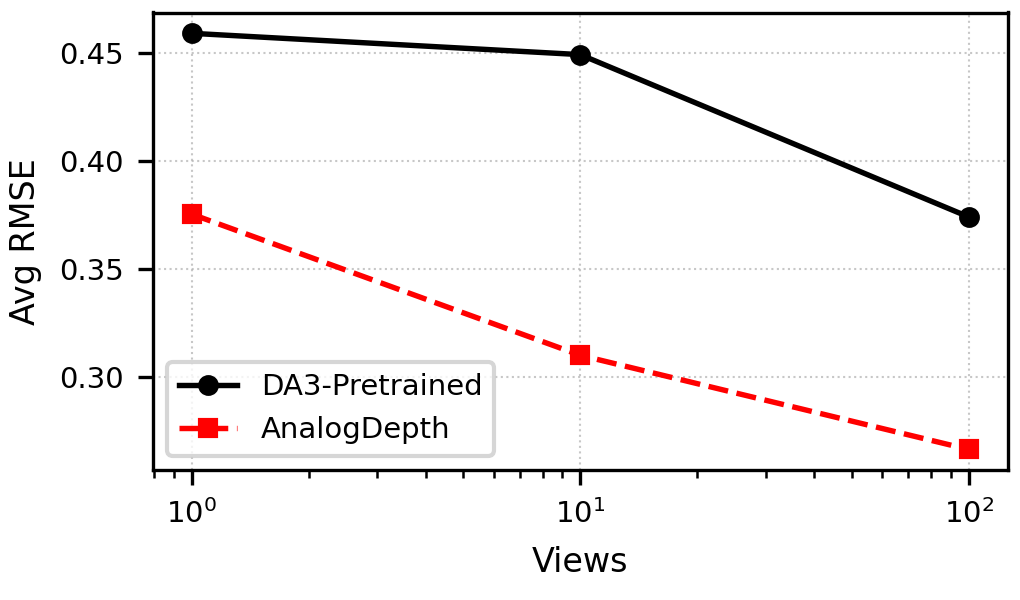}
  \caption{Average Depth RMSE vs Number of images used in inference. RMSE is the average of all frames sampled across the test-set.}
  \label{fig:rmse_vs_views}
\end{figure}

\begin{figure}[!tb]
    \centering

    % Row 1
    \includegraphics[width=0.325\linewidth]{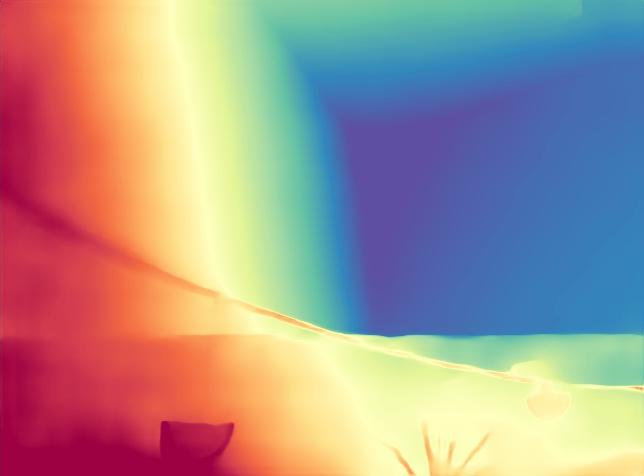}
    \includegraphics[width=0.325\linewidth]{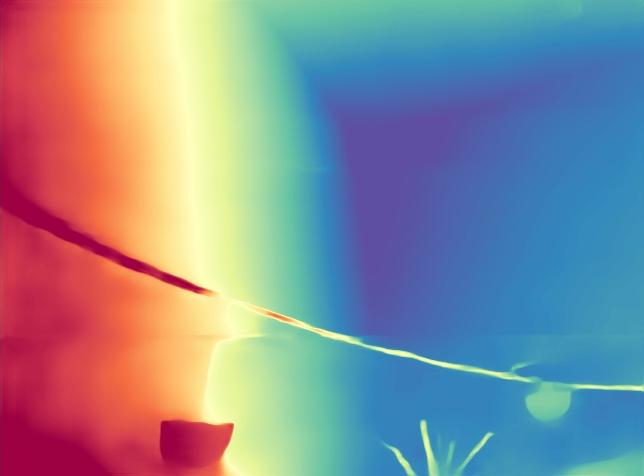}
    \includegraphics[width=0.325\linewidth]{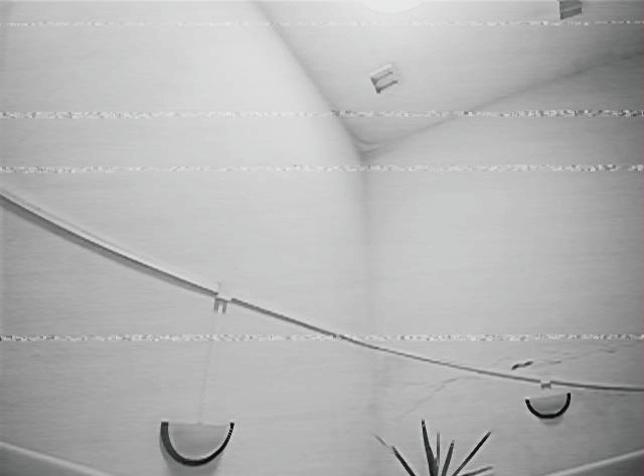}

    \vspace{2mm}

    % Row 2
    \includegraphics[width=0.325\linewidth]{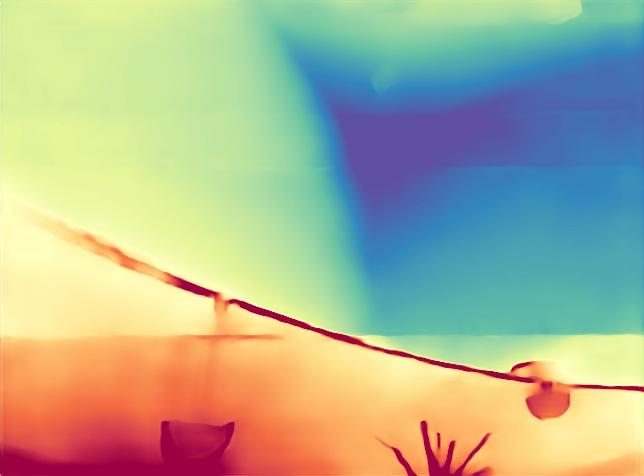}
    \includegraphics[width=0.325\linewidth]{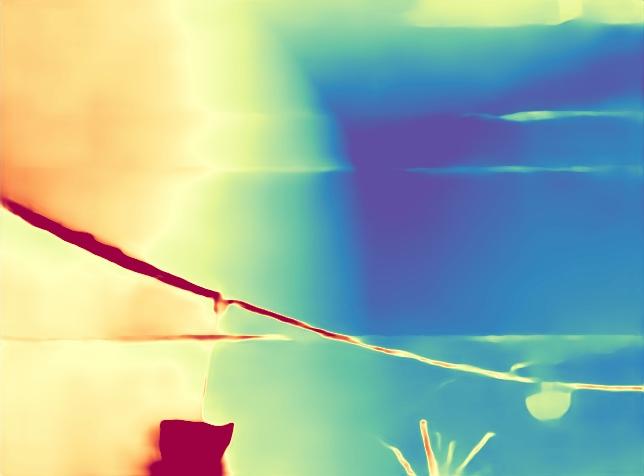}
    \includegraphics[width=0.325\linewidth]{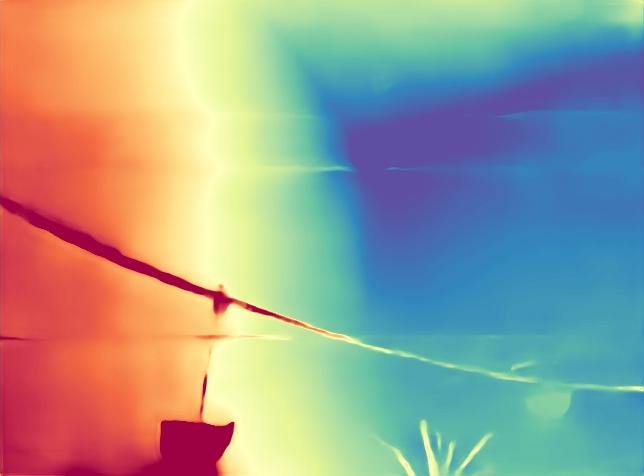}

        % Text row
    % \makebox[0.31\columnwidth]{Text 1}
     \makebox[0.31\columnwidth]{{\small 1 view }}
    \makebox[0.31\columnwidth]{{\small 10 views}}
    \makebox[0.31\columnwidth]{{\small 100 views}}

    \caption{Depth estimation for one corrupted analog VTX image by varying the number of images passed to AnalogDepth (\textit{top}) and DA3-pretrained (\textit{Bottom}).}
    \label{fig:view_ablation}
\end{figure}

\subsection{Multi-view Ablation}

Increasing the number of views in the cross-attention should improve depth accuracy since analog VTX noise is transient. We tested, in Fig \ref{fig:rmse_vs_views}, if increasing the input of number of views makes our model redundant compared to DA3 pretrained. By increasing the number of images from single-image to a total of 100 images the RMSE drops for both models. This indicates that increasing the number of views is not enough. In Fig. \ref{fig:view_ablation} we can see that AnalogDepth with Single Image is affected by the horizontal bands however the structure is recovered when we use multi-view with 10 images. On the other hand for pretrained DA3 10 views were not enough to remove artifacts. Even with 100 it still shows horizontal lines. In terms of runtime, inference took on a GeForce RTX 2080 Ti respectively: 0.05 s for a single image, 0.35 s for 10 images and 7.12 s for 100 images.

\subsection{Reconstruction Results}

\begin{figure*}[!tb]
    \centering

    % Row 1
    \includegraphics[width=0.23\linewidth]{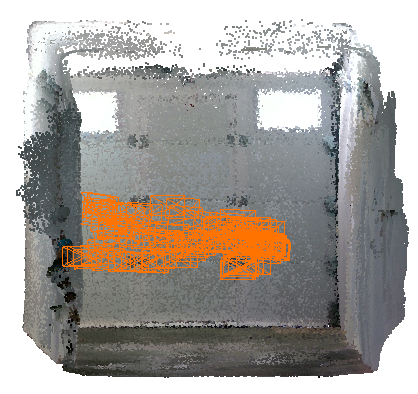}
    \includegraphics[width=0.23\linewidth]{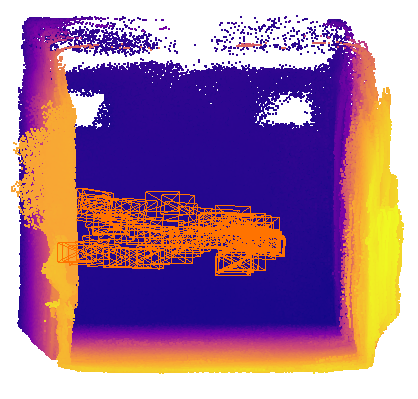}
    \includegraphics[width=0.23\linewidth]{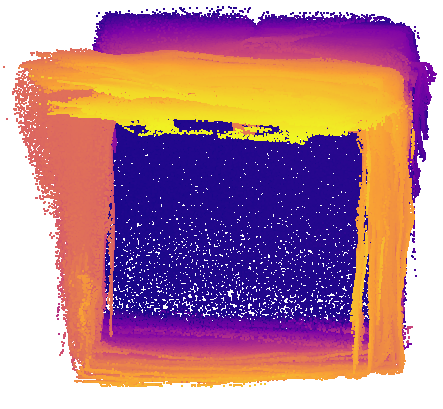}
    \includegraphics[width=0.245\linewidth]{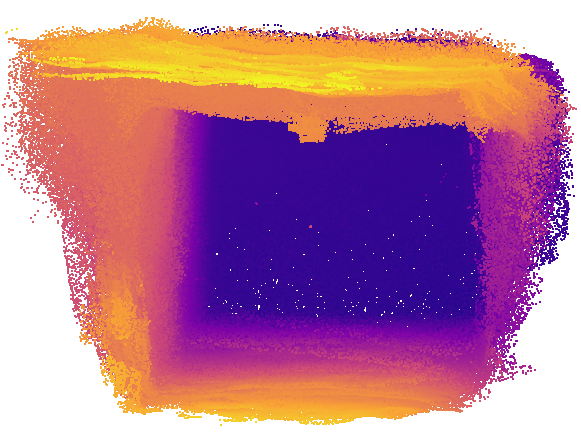}
    \vspace{2mm}
     \includegraphics[width=0.245\linewidth]{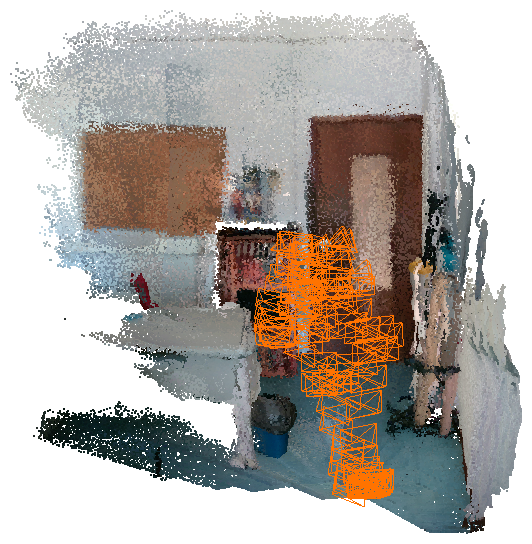}
    \includegraphics[width=0.245\linewidth]{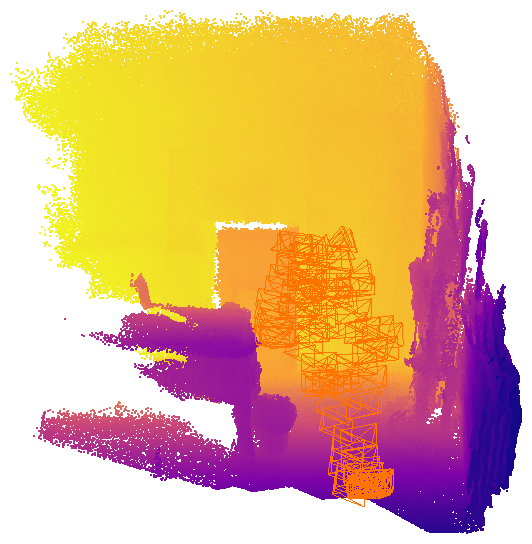}
    \includegraphics[width=0.21\linewidth]{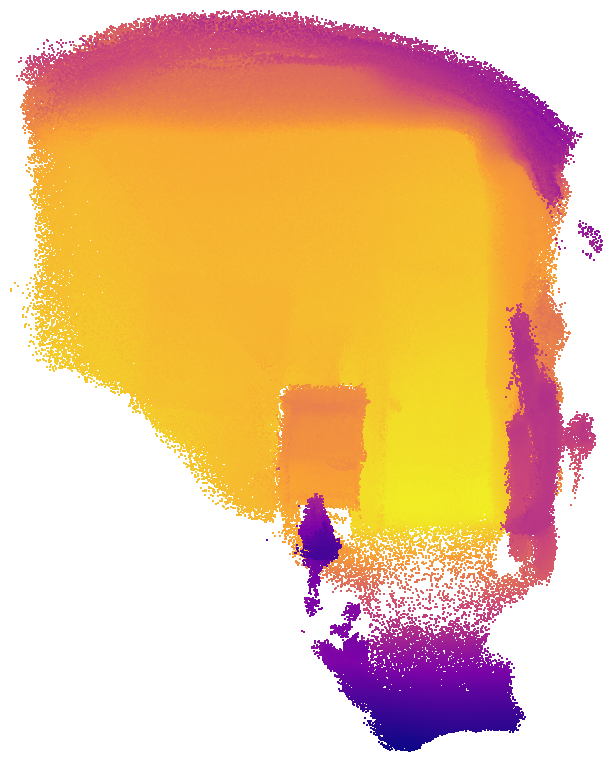}
    \includegraphics[width=0.23\linewidth]{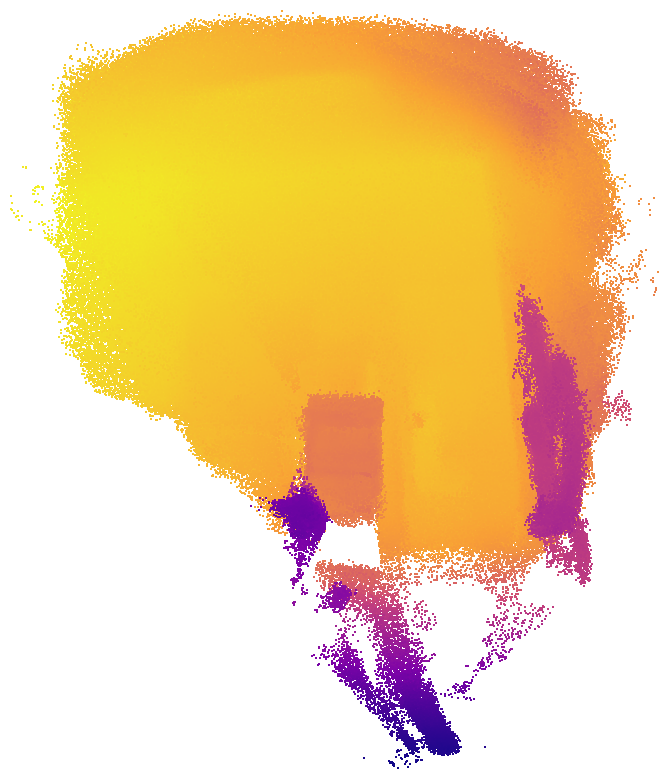}

    \includegraphics[width=0.245\linewidth]{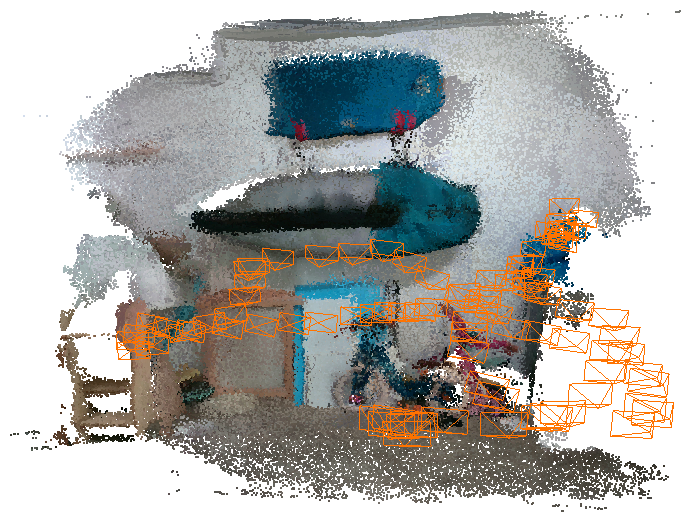}
    \includegraphics[width=0.235\linewidth]{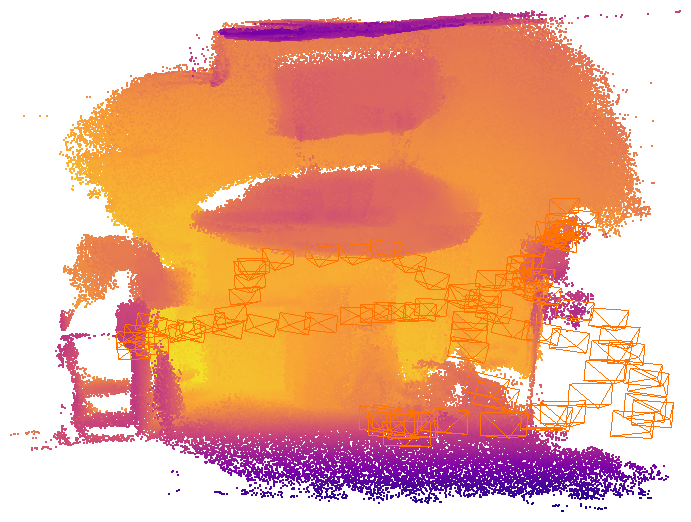}
    \includegraphics[width=0.245\linewidth]{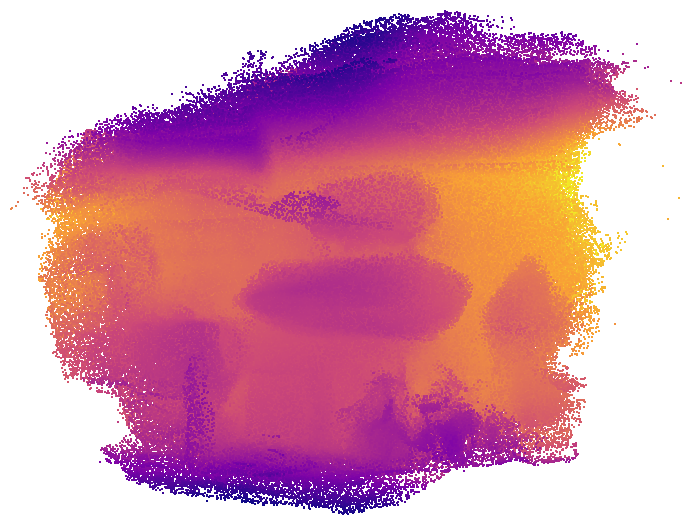}
    \includegraphics[width=0.245\linewidth]{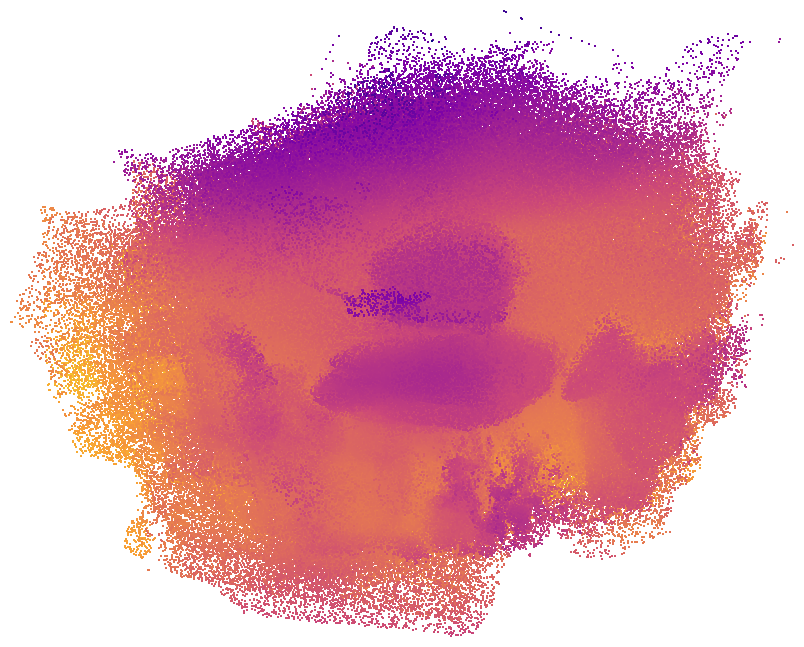}

    \makebox[0.245\textwidth]{{\small RealSense-DA3 RGB}}
    \makebox[0.245\textwidth]{{\small RealSense-DA3 }}
    \makebox[0.245\textwidth]{{\small AnalogDepth}}
    \makebox[0.245\textwidth]{{\small DA3-pretrained}}

    \caption{Reconstructions for our test-set scenes (top to bottom): garage\_west, office and  garage\_south. Rendered point clouds are colored by Z buffer. (\textit{Top}) The right wall on DA3-pretrained appears smeared and distorted. (\textit{Middle}) The walls are less defined and the ceiling is in the same depth plane as the back wall. (\textit{Bottom}) The scene looks flat for the DA3-pretrained compared to AnalogDepth. Notice the floor and ceiling.}
    \label{fig:3D_render}
\end{figure*}

As shown on Table \ref{tab:chamfer}, Similarly to the depth map accuracy, training with a noise bank improved the reconstruction for all sequences except the \textit{office\_1}. We observed that for that flight, the reconstructions have very little overlap. In the \textit{garage\_west} reconstructions, shown in Fig. \ref{fig:3D_render}, it is especially clear that DA3-pretrained distorts consistently the right wall causing a smear of points. Similar conclusions can be drawn from the other reconstructions where room orthogonality and planar structure are less well preserved.

\begin{table}[!tb]
  \centering
  \setlength{\tabcolsep}{5pt}
  \renewcommand{\arraystretch}{1.3}
  \begin{tabular}{lcccc}
  \toprule
  Video & Pretrained & Noise Bank \\
  \midrule
  garage\_west& 0.567 & \textbf{0.495}  \\
  garage\_south & 0.407  & \textbf{0.253} \\
  office\_1 & \textbf{0.841} & 1.000 \\
  office\_2 & 0.456 & \textbf{0.389} \\
  plants & 3.281 & \textbf{2.186} \\
  stairs & 1.112 & \textbf{0.776} \\
  \bottomrule
  \end{tabular}
  \caption{Reconstruction Error: Chamfer Distance between aligned point clouds}
  \label{tab:chamfer}
  \end{table}

%%%%%%%%%%%%%%%%%%%%%%%%%%%%%%%%%%%%%%%%%%%%%%%%%%%%%%%%%%%%%%%%%%%%%%%%%%%%%%%%
\section{CONCLUSION}

We have presented AnalogDepth, a method for adapting the DA3 visual geometry foundation model to the degraded imagery produced by analog FPV video transmission. By analyzing real analog VTX noise through its power spectral density, we showed that it exhibits: spatial structure, low-frequency energy concentration and strong vertical banding that cannot be replicated by AWGN or PSD-matched Gaussian synthesis alone. Using a student-teacher knowledge distillation strategy with LoRA, we efficiently fine-tune the DINOv2 backbone of DA3 with noise sampled from a real analog noise bank, without requiring access to the original training procedure or datasets.

Experiments on six real FPV flight sequences across three indoor scenes demonstrate consistent improvements in both per-frame depth RMSE and 3D reconstruction Chamfer distance compared to the pretrained DA3 baseline and both Gaussian noise variants. A multi-view ablation confirms that simply increasing the number of input views does not compensate for the domain gap introduced by analog transmission, motivating the need for dedicated adaptation. The results were also consistent for different ground-truth constructions. Qualitatively, AnalogDepth recovers accurate planar geometry on textureless walls where the pretrained model systematically fails.

Future work will explore larger and more diverse noise banks collected across environments and interference conditions, extension to outdoor FPV sequences and different VTX hardware setups, and joint adaptation of both the backbone and the DA3 decoder heads for further accuracy gains. Furthermore, we will extend our work to other visual geometry models, including models that include Metric Scale.

%%%%%%%%%%%%%%%%%%%%%%%%%%%%%%%%%%%%%%%%%%%%%%%%%%%%%%%%%%%%%%%%%%%%%%%%%%%%%%%%

\bibliographystyle{IEEEtran}
\bibliography{refs}

@inproceedings{yang2024depthanythingv2,
  author    = {Yang, Lihe and Kang, Bingyi and Huang, Zilong and Zhao, Zhen and Xu, Xiaogang and Feng, Jiashi and Zhao, Hengshuang},
  title     = {Depth Anything {V2}},
  booktitle = {Advances in Neural Information Processing Systems (NeurIPS)},
  year      = {2024},
}

@inproceedings{wang2024dust3r,
  author    = {Wang, Shuzhe and Leroy, Vincent and Cabon, Yohann and Chidlovskii, Boris and Revaud, J{\'e}r{\^o}me},
  title     = {{DUSt3R}: Geometric 3{D} Vision Made Easy},
  booktitle = {Proc. IEEE/CVF Conf. Computer Vision and Pattern Recognition (CVPR)},
  year      = {2024},
}

@inproceedings{leroy2024mast3r,
  author    = {Leroy, Vincent and Cabon, Yohann and Revaud, J{\'e}r{\^o}me},
  title     = {Grounding Image Matching in 3{D} with {MASt3R}},
  booktitle = {Proc. European Conf. Computer Vision (ECCV)},
  year      = {2024},
}

@article{bochkovskii2024depthpro,
  author    = {Bochkovskii, Aleksei and Dellenbach, Ama{\"e}l and Richter, Hugo and Bhat, Shariq Farooq and Yang, Yichao and Koltun, Vladlen},
  title     = {Depth {Pro}: Sharp Monocular Metric Depth in Less Than a Second},
  journal   = {arXiv preprint arXiv:2410.02073},
  year      = {2024},
}

@inproceedings{wang2025vggt,
  author    = {Wang, Jianyuan and Chen, Minghao and Karaev, Nikita and Vedaldi, Andrea and Rupprecht, Christian and Novotny, David},
  title     = {{VGGT}: Visual Geometry Grounded Transformer},
  booktitle = {Proc. IEEE/CVF Conf. Computer Vision and Pattern Recognition (CVPR)},
  year      = {2025},
}

@article{lin2025depthanything3,
  title={Depth anything 3: Recovering the visual space from any views},
  author={Lin, Haotong and Chen, Sili and Liew, Junhao and Chen, Donny Y and Li, Zhenyu and Shi, Guang and Feng, Jiashi and Kang, Bingyi},
  journal={arXiv preprint arXiv:2511.10647},
  year={2025}
}

@inproceedings{kong2023robodepth,
  author    = {Kong, Lingdong and Xie, Shaoyuan and Hu, Hanjiang and Cottereau, Benoit R. and others},
  title     = {{RoboDepth}: Robust Out-of-Distribution Depth Estimation under Corruptions},
  booktitle = {Advances in Neural Information Processing Systems (NeurIPS)},
  year      = {2023},
}

@inproceedings{zamir2022restormer,
  author    = {Zamir, Syed Waqas and Arora, Aditya and Khan, Salman and Hayat, Munawar and Khan, Fahad Shahbaz and Yang, Ming-Hsuan},
  title     = {Restormer: Efficient Transformer for High-Resolution Image Restoration},
  booktitle = {Proc. IEEE/CVF Conf. Computer Vision and Pattern Recognition (CVPR)},
  year      = {2022},
}

@inproceedings{chen2022nafnet,
  author    = {Chen, Liangyu and Chu, Xiaojie and Zhang, Xiangyu and Sun, Jian},
  title     = {Simple Baselines for Image Restoration},
  booktitle = {Proc. European Conf. Computer Vision (ECCV)},
  year      = {2022},
}

@inproceedings{potlapalli2023promptir,
  author    = {Potlapalli, Vaishnav and Zamir, Syed Waqas and Khan, Salman and Khan, Fahad Shahbaz},
  title     = {{PromptIR}: Prompting for All-in-One Blind Image Restoration},
  booktitle = {Advances in Neural Information Processing Systems (NeurIPS)},
  year      = {2023},
}

@inproceedings{wang2020tartanair,
  title={Tartanair: A dataset to push the limits of visual slam},
  author={Wang, Wenshan and Zhu, Delong and Wang, Xiangwei and Hu, Yaoyu and Qiu, Yuheng and Wang, Chen and Hu, Yafei and Kapoor, Ashish and Scherer, Sebastian},
  booktitle={2020 IEEE/RSJ International Conference on Intelligent Robots and Systems (IROS)},
  pages={4909--4916},
  year={2020},
  organization={IEEE}
}

@inproceedings{hu2021lora,
  author    = {Hu, Edward J. and Shen, Yelong and Wallis, Phillip and Allen-Zhu, Zeyuan and Li, Yuanzhi and Wang, Shean and Wang, Lu and Chen, Weizhu},
  title     = {{LoRA}: Low-Rank Adaptation of Large Language Models},
  booktitle = {Proc. Int. Conf. Learning Representations (ICLR)},
  year      = {2022},
}

@inproceedings{lindenberger2023lightglue,
  author    = {Lindenberger, Philipp and Sarlin, Paul-Edouard and Pollefeys, Marc},
  title     = {{LightGlue}: Local Feature Matching at Light Speed},
  booktitle = {Proc. IEEE/CVF Int. Conf. Computer Vision (ICCV)},
  year      = {2023},
}

@inproceedings{fan2017pointcloud,
  author    = {Fan, Haoqiang and Su, Hao and Guibas, Leonidas},
  title     = {A Point Set Generation Network for 3{D} Object Reconstruction from a Single Image},
  booktitle = {Proc. IEEE/CVF Conf. Computer Vision and Pattern Recognition (CVPR)},
  year      = {2017},
}

@article{kasdin1995,
  author    = {Kasdin, N. Jeremy},
  title     = {Discrete Simulation of Colored Noise and Stochastic Processes and $1/f^\alpha$ Power Law Noise Generation},
  journal   = {Proc. IEEE},
  volume    = {83},
  number    = {5},
  pages     = {802--827},
  year      = {1995},
}

@article{dao2023flashattention2,
    title   = {{FlashAttention-2}: Faster Attention with Better Parallelism and Work Partitioning},
    author  = {Dao, Tri},
    journal = {arXiv preprint arXiv:2307.08691},
    year    = {2023}
  }

@article{proencca2018probabilistic,
  title={Probabilistic RGB-D odometry based on points, lines and planes under depth uncertainty},
  author={Proen{\c{c}}a, Pedro F and Gao, Yang},
  journal={Robotics and Autonomous Systems},
  volume={104},
  pages={25--39},
  year={2018},
  publisher={Elsevier}
}

@inproceedings{gasperini2023robust,
  title={Robust monocular depth estimation under challenging conditions},
  author={Gasperini, Stefano and Morbitzer, Nils and Jung, HyunJun and Navab, Nassir and Tombari, Federico},
  booktitle={2023 IEEE/CVF International Conference on Computer Vision (ICCV)},
  pages={8143--8152},
  year={2023},
  organization={IEEE}
}

@inproceedings{jiang2025always,
  title={Always Clear Depth: Robust Monocular Depth Estimation Under Adverse Weather.},
  author={Jiang, Kui and Cao, Jing and Yu, Zhaocheng and Jiang, Junjun and Zhou, Jingchun},
  booktitle={IJCAI},
  pages={1251--1259},
  year={2025}
}

\end{document}